\documentclass[11pt,a4paper]{article}
\usepackage[hyperref]{ranlp2023}
\aclfinalcopy

\usepackage{times}
\usepackage{latexsym}
\usepackage{graphicx}
\usepackage{booktabs}

\usepackage[skins]{tcolorbox}
\usepackage[toc,page]{appendix}
\usepackage{microtype}
\usepackage{multirow}
\aclfinalcopy 

\title{Using LLMs for Multilingual Clinical Entity Linking to ICD-10}

\author{Sylvia Vassileva \\
  Faculty of Mathematics and Informatics,\\ Sofia University St. Kliment Ohridski\\
  \texttt{svasileva@fmi.uni-sofia.bg} \\\And
  Ivan Koychev \\
  Faculty of Mathematics and Informatics,\\ Sofia University St. Kliment Ohridski\\
  \texttt{koychev@fmi.uni-sofia.bg} \\\AND
  Svetla Boytcheva \\
  Graphwise \& \\ Faculty of Mathematics and Informatics,\\ Sofia University St. Kliment Ohridski \\  
  \texttt{svetla.boytcheva@graphwise.ai} \\}

\date{}

\begin{document}
\maketitle
\begin{abstract}
The linking of clinical entities is a crucial part of extracting structured information from clinical texts. It is the process of assigning a code from a medical ontology or classification to a phrase in the text. The International Classification of Diseases - 10th revision (ICD-10) is an international standard for classifying diseases for statistical and insurance purposes. Automatically assigning the correct ICD-10 code to terms in discharge summaries will simplify the work of healthcare professionals and ensure consistent coding in hospitals. 
Our paper proposes an approach for linking clinical terms to ICD-10 codes in different languages using Large Language Models (LLMs). The approach consists of a multistage pipeline that uses clinical dictionaries to match unambiguous terms in the text and then applies in-context learning with GPT-4.1 to predict the ICD-10 code for the terms that do not match the dictionary. Our system shows promising results in predicting ICD-10 codes on different benchmark datasets in Spanish - 0.89 F1 for categories and 0.78 F1 on subcategories on CodiEsp, and Greek - 0.85 F1 on ElCardioCC.
\end{abstract}

\section{Introduction}

Medical coding and entity linking to standard classifications are very important tasks in the domains of healthcare management and research. One of the widely used standard classifications is the International Statistical Classification of Diseases and Related Health Problems - 10th Revision (ICD-10) \footnote{\url{https://icd.who.int/browse10/2019/en}} -  translated into 40+ languages and used in 100+ countries.

Named Entity Recognition (NER) in clinical text is highly complex, especially in multilingual settings. The main challenges are the lack of annotated datasets, domain-specific linguistic tools, and natural language processing (NLP) models. Additionally, clinical data is sensitive, limiting access and use, and creating annotated corpora is time-consuming and produces only small datasets. These factors set some constraints in the development of supervised NLP models for linking clinical concepts to ICD-10 codes.

In this paper, we present a fully unsupervised approach for entity linking of clinical texts in Spanish and Greek to ICD-10. The proposed solution incorporates a dictionary-based approach, which ensures high precision and large language models (LLMs) that provide robustness and variability of paraphrases, concepts out of vocabulary, and context-dependent mappings in case of ambiguous mentions. For evaluation of the proposed approach, we selected two benchmark datasets ElcardioCC\footnote{\url{https://elcardiocc.web.auth.gr/}}, consisting of discharge summaries in Greek, and CodiESP\footnote{\url{https://temu.bsc.es/codiesp/}} with clinical texts in Spanish. The achieved results are very promising and show that our system outperforms most of the supervised models for the same benchmark datasets \footnote{The code and prompts are available on our \href{https://github.com/svassileva/llm_multilingual_linking_icd10}{GitHub}}.

\section{Related Work}

The full spectrum of supervised, unsupervised, and hybrid approaches has been explored for multilingual clinical text entity linking with ICD-10.
For Spanish clinical texts, supervised approaches employing BERT achieved mean average precision (MAP) scores of 0.482 \cite{lopez2020icb} and 0.517 \cite{costa2020fraunhofer}, as well as an F1 score of 0.505 using CRF (Conditional Random Field) and BERT (fine-tuned). In \citet{garcia2020fle}, a hybrid approach was proposed using BERT-based NER, semantic linking, text augmentation, and a knowledge graph, achieving an F1-score of 0.679 for the CodiEsp dataset. In French, English, and Japanese scenarios, hybrid approaches blending dictionary projection, rule-based methodologies, neural networks, and retrieval techniques, achieved F1 scores ranging from 0.694 up to 0.8586 and accuracy in the range 0.75-0.86 (\citet{seva2017multi}, \citet{Zweigenbaum2016Hybrid}, \citet{miftahutdinov2018deep}, \citet{Sheng2014NCU}). In \citet{Henning2020Multilingual}, a multilingual ensemble method was proposed that incorporated BioBERT, ClinicalBERT, XLNet, GEMs (General Equivalence Mappings), and achieved MAP: 0.259 (overall), 0.306 (subset), and F1 score 0.608 for MIMIC III Top 50.

With the emergence of LLM models, several studies investigated their capabilities for the ICD-10 coding task \cite{pathak2024utilizing}. The first attempts did not present promising results, but when more mature GenAI technologies became available and expertise in prompt engineering was gained, the results also improved - current publications report results comparable with supervised models \cite{mustafa2025large}, achieving accuracy from 0.86 to 0.89. The authors of \citet{li2024exploring} reported results for MIMIC-III top 50 Codes dataset with GPT-4 (5-shot) that achieved Micro-F1 0.589. \citet{soroush2023assessing} investigated the performance of GPT-3.5 and GPT-4 for ICD-10 coding and reported exact match for codes 0.13 and billable codes 0.77. 
\citet{boyle2023automated} presented results of LLMs ICD-10 coding for CodiEsp Spanish dataset comparing GPT-3.5, GPT-4 and Llama-2, achieving macro F1-score 0.225 for GPT-4. Another study for the CodiEsp-X task and GPT-4 \cite{puts2025developing} reported an F1-score of 0.305. LLMs for ICD-10 coding in Swedish were explored in \citet{karlin2025prompt}, achieving micro F1-scores of 0.279 and 0.126 for LlaMa-3.1 and GPT-SW3-6.7B, respectively. In \citet{maatouk2025leveraging}, T5 models for ICD-10 coding were evaluated and have shown a micro F1-score of 0.487.

\section{Data}
\textbf{Spanish Clinical Dataset - CodiEsp} 
CodiEsp \cite{CLEFeHealth2020Task1Overview} is a Spanish clinical dataset consisting of 1,000 patient discharge summaries labeled by medical professionals. In each discharge summary, the experts labeled all diagnoses and procedures as spans in the text and assigned the best ICD-10-CM or ICD-10-PCS codes, respectively.
In this paper, we use only the diagnoses part of the CodiEsp dataset.

\textbf{Greek Clinical Dataset - ElCardioCC}
The ELCardioCC dataset \footnote{\url{https://elcardiocc.web.auth.gr/}} \cite{BioASQ2025ElCardioCC} consists of 1,000 de-identified hospital discharge letters in Greek written by cardiology doctors. The dataset was labeled by medical professionals who identified all spans in the text related to chief complaint, diagnosis, prior medical history, and findings. Each span was assigned an ICD-10 code based on the term's meaning in the context. 

\textbf{ICD-10 Dictionaries}
We compiled dictionaries with clinical terms and their ICD-10 codes using the following resources: ICD-10 specifications in Spanish (CIE-10)\footnote{\url{https://www.sanidad.gob.es/en/estadEstudios/estadisticas/normalizacion/clasifEnferm/home.htm}} and Greek \footnote{\url{https://medicalcodes.instdrg.gr/search/icd/systematic}}; CodeEsp train set terms - Spanish; ElCardioCC train set terms - Greek.
The Spanish dictionary contains about 88K terms and their ICD-10 mappings (up to 4 characters), and the Greek dictionary - about 11,500 terms and their 3-character ICD-10 codes.

\section{Methods}
 \begin{figure*}[h]
	\centering
	\includegraphics[width=0.9\textwidth]{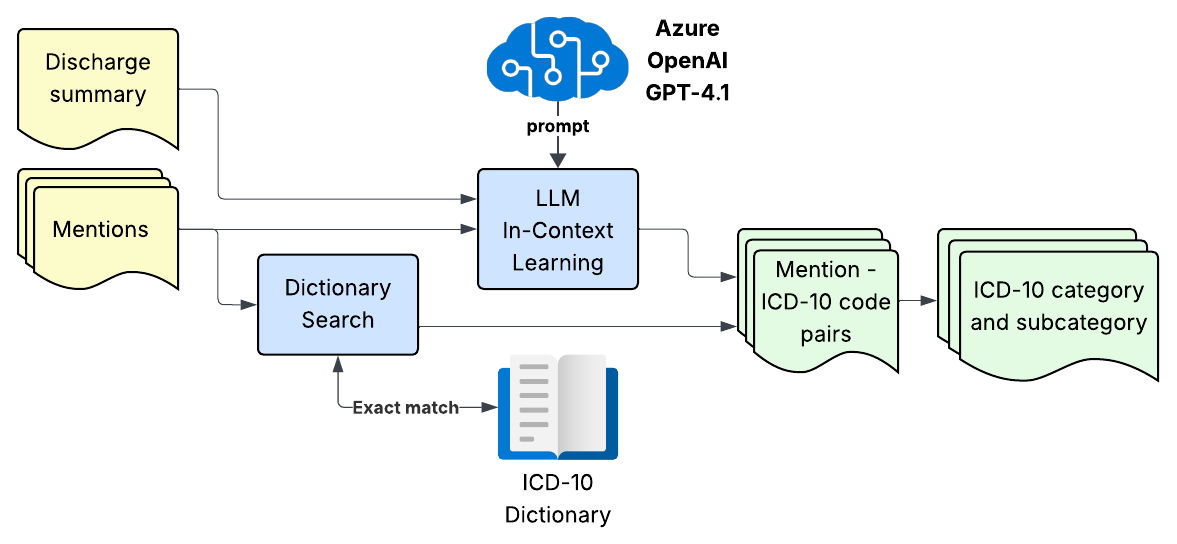}
	\caption{The process for entity linking uses multiple stages - first, each mention in the text is matched against an unambiguous ICD-10 dictionary in the specific language, second, an LLM is prompted to select the best ICD-10 for all mentions in the discharge summary, and finally, the results are combined.}
	\label{fig:method}
\end{figure*}

We focus on the task of entity linking to ICD-10, which consists of assigning the most appropriate code from about 2K categories or 14K subcategories. The task is an extreme multi-class classification where multiple codes can apply to a term, but one is assigned based on the patient context.
Our system accepts patient term mentions and discharge summaries and assigns ICD-10 codes to each mention. A mention is a span in a sentence that represents a concept from ICD-10. For example, in the sentence \textit{"The patient presented with a headache."}, the term \textit{"headache"} is a mention of the concept with code R51. Our pipeline addresses only the entity linking of the mention. Detecting the mention is also a challenging task that should be addressed as future work.
The process consists of several steps - first, we search in the language-specific dictionaries of ICD-10 codes to find an exact match for the mention. If we find a match which corresponds unambiguously to a specific ICD-10 code, we return that code. Otherwise, the mention is assigned a code at the second step.
In the second step, we use an LLM (GPT-4.1) and prompt it to generate ICD-10 codes for all mentions in the text as well as an explanation for each assignment. We use in-context learning and provide one example of a discharge summary in the same language with labeled mentions and respective ICD-10 codes. We generated the example using GPT-4o. 
The process of ICD-10 code assignment is presented in Figure \ref{fig:method}.
Finally, we process the results from both steps and assign an ICD-10 code if the dictionary match or LLM response returns a result for the mention.
The proposed approach can be easily adapted to other languages using ICD-10 specification in the specific language and taking advantage of the multilingual support of GPT-4 models. The prompts we utilized follow the same structure for both languages, with parameters to specify the language and language-specific examples generated by an LLM.
In our experiments, we compare 3- and 4-character codes if they are available in the dataset.

\begin{table*}[h]
\begin{center}
\begin{tabular}{@{}lrrrrrrrrr@{}}
\toprule
\multicolumn{1}{c}{\textbf{Model}} & \multicolumn{3}{c}{\textbf{Greek}} & \multicolumn{3}{c}{\textbf{Spanish Cat}} & \multicolumn{3}{c}{\textbf{Spanish Subcat}}\\
\midrule
 & 
\textbf{P} & 
\textbf{R} & 
\textbf{F1} &
\textbf{P} & 
\textbf{R} & 
\textbf{F1} &
\textbf{P} & 
\textbf{R} & 
\textbf{F1} 
\\ \midrule
Dict	 &	0.657 &	0.657 & 0.657 & 0.546 &	0.546 & 0.546 & 0.528 &	0.528 & 0.528 \\
GPT-4o 0-shot	 &	0.814	& 0.462	&0.589 &	0.782 &	0.542&	0.640 &	0.641 &	0.445&	0.525\\
Dict+GPT-4o 0-shot	 &	0.776	&0.776&	0.776 &	0.776 &	0.776 & 0.776  &	0.641 &	0.445&	0.525\\
GPT-4o 1-shot	 &	0.823 &	0.324 & 0.465  &	0.810	 &0.113	 &0.199 &	0.674	 &0.094	 &0.166 \\
Dict+GPT-4o 1-shot	 &	0.730 &	0.730 & 0.730 &	0.597	 &0.597 &	0.597 &	0.5966	 &0.597 &	0.597\\
GPT-4.1 0-shot &	0.822 &	0.808 &	0.815 &	0.774	& 0.774& 	0.774 & 0.566	 &0.566	 &0.566 \\
Dict+GPT-4.1 0-shot	 &	\underline{0.853}	 &\underline{0.853}	 &\underline{0.853} 	& \underline{0.890} &	\underline{0.890}& 	\underline{0.890}	& \textbf{0.780}& \textbf{0.780}	& \textbf{0.780} \\
GPT-4.1 1-shot	 &	0.823 &	0.822 & 0.823 &	0.775	&0.773&	0.774  &		0.610 &	0.609&	0.609\\
Dict+GPT-4.1 1-shot &	\textbf{0.856} &	\textbf{0.856} & \textbf{0.856} &	\textbf{0.891}	&\textbf{0.891} &	\textbf{0.891} &	\underline{0.776}&	\underline{0.776} &	\underline{0.776}\\
\bottomrule
\end{tabular}
\end{center}
\caption{Results of evaluation of models on the ElCardioCC dataset in Greek on the category level and the CodiEsp dataset (diagnoses) in Spanish on the category and subcategory levels. The highest score for each metric is shown in bold, and the second highest is underlined.}
\label{tab:entity_linking_results}
\end{table*}

\section{Experiments and Results}
As a baseline method, we used the dictionary exact match approach. We perform experiments with two LLMs - GPT-4o and GPT-4.1, using a private Azure deployment in order to protect patient data.
We used a temperature of 0.5 and 6K max tokens in order to accommodate the length of discharge summary texts.
We compare zero- and one-shot prompts with both models, as well as combining the dictionary approach with the LLM. We limited our experiments to one shot, as adding more than one example discharge summary and its mentions increases the token count significantly.

The results of our experiments on the datasets are shown in Table \ref{tab:entity_linking_results}. On the Greek dataset, GPT-4o was not able to outperform the dictionary baseline. However, combining both dictionary and GPT-4o zero- or one-shot improved compared to the baseline F1 with 0.12 and 0.07, respectively. Zero-shot outperforms one-shot, which is unusual but may be explained by the lengthy prompt and the inability of the model to generate codes for all mentions. 
GPT-4.1 improves the result compared to the baseline, and zero- and one-shot scores are quite close, with one-shot performing a bit better (0.82 F1). Combining GPT-4.1 one-shot with the dictionary shows the best result on the Greek dataset - 0.85 F1, followed by GPT-4.1 zero-shot and the dictionary.

For the Spanish dataset, we also compare the performance of the model on category and subcategory levels on the diagnoses part of CodiEsp, since the dataset is labeled with more specific codes. Predicting the correct category is an easier task, so the F1 scores are higher.

Similarly to the Greek dataset experiments, we observe that GPT-4.1 performs better than GPT-4o and in some cases, the differences are very significant. The best model combines the dictionary approach and GPT-4.1 one-shot and achieves 0.8906 F1 on the CodiEsp test dataset. Again, the second-best model is the zero-shot GPT 4.1, including dictionary predictions - F1 score 0.8902. 

The only difference in subcategory evaluation is that the GPT-4.1 zero-shot performed slightly better than the one-shot, and combined with the dictionary method, achieved 0.78 F1, while the one-shot - 0.77 F1. GPT-4o showed consistently lower results, especially due to low recall.

In general, GPT-4o shows relatively good precision but very poor recall, due to the fact that it does not generate codes for all mentions. GPT-4.1, on the other hand, shows balanced precision and recall and performs the best overall in both languages. Adding the dictionary to the system improves the overall F1 score.

\subsection{Discussion}
The main challenge in our LLM experiments was to optimize the prompt so that the LLM retrieves the codes for as many mentions as possible. This was a particular problem with GPT-4o and is the main factor contributing to its lower performance. The length of the discharge summary and terms context is also a challenge, even though the OpenAI models support very long contexts.
Another challenge specific to ICD-10 coding is the classification of symptoms pointing to a specific diagnosis or not classified elsewhere (falling under the R00-R99 chapter). Since this classification requires ruling out all other categories, LLMs struggle to correctly identify it, unless the term is a really common symptom like \textit{"headache"}.

\subsection{Limitations}
Our experiments were performed with two different languages - Spanish and Greek, due to the challenge of finding clinical datasets with ICD-10 labels. Therefore, applying the approach in different languages may not show the same results as the availability of ICD-10 dictionaries, and the performance of GPT-4.1 may vary based on language.
Furthermore, the experiments were performed only with Azure OpenAI LLMs, making it impossible to run the system on hospital premises, which may be a limitation for adoption in some environments. Also, the cost of these models for inference is higher than using smaller open-source models, which can be a limiting factor for usage.

\section{Conclusion}
In this paper, we presented an approach for multilingual entity linking to ICD-10, which will help automate the assignment of codes in hospitals and offload some of the coding burden from medical professionals. Our approach shows promising results in two very different languages - Spanish and Greek, and does not require specific training data, only ICD-10 term dictionaries. The system using dictionaries and GPT-4.1 achieved a 0.85 F1 score on the Greek dataset, and 0.89 F1 / 0.78 F1 on the Spanish ICD-10 categories and subcategories, respectively. This demonstrates the potential of LLMs to help with medical ICD-10 coding without further fine-tuning on labeled datasets.
As future work, experimenting with fine-tuning open-source and smaller LLMs would be beneficial, since they can be hosted locally on hospital premises and be more efficient and less expensive. 
Our experiments focused on entity linking, specifically, however, the process of detecting terms in the text can result in mentions with incorrect boundaries, missing or extra mentions, which will also increase the errors in the linking. Therefore, it would be useful to explore the end-to-end detection and linking process in the future and experiment with LLMs to address this problem.

\section*{Acknowledgments}
This work was partially supported by the European Union-NextGenerationEU, through the National Recovery and Resilience Plan of the Republic of Bulgaria [Grant Project No. BG-RRP-2.004-0008].
Part of this work is also supported by European Union’s Horizon research and innovation programme projects RES-Q PLUS [Grant Agreement No. 101057603] and HEREDITARY [Grant Agreement No. 101137074]. Views and opinions expressed are however, those of the author only and do not necessarily reflect those of the European Union. Neither the European Union nor the granting authority can be held responsible for them.

\bibliographystyle{acl_natbib}
\bibliography{paper}

\begin{thebibliography}{18}
\expandafter\ifx\csname natexlab\endcsname\relax\def\natexlab#1{#1}\fi

\bibitem[{Boyle et~al.(2023)Boyle, Kascenas, Lok, Liakata, and O'Neil}]{boyle2023automated}
Joseph~S Boyle, Antanas Kascenas, Pat Lok, Maria Liakata, and Alison~Q O'Neil. 2023.
\newblock Automated clinical coding using off-the-shelf large language models.
\newblock \emph{arXiv preprint arXiv:2310.06552}.

\bibitem[{Costa et~al.(2020)Costa, Lopes, Carreiro, Ribeiro, and Soares}]{costa2020fraunhofer}
Joao Costa, In{\^e}s Lopes, Andr{\'e}~V Carreiro, David Ribeiro, and Carlos Soares. 2020.
\newblock \href {https://ceur-ws.org/Vol-2696/paper_187.pdf} {Fraunhofer aicos at clef ehealth 2020 task 1: Clinical code extraction from textual data using fine-tuned bert models.}
\newblock In \emph{CLEF (Working Notes)}.

\bibitem[{Dimitriadis et~al.(2025)Dimitriadis, Patsiou, Stoikopoulou, Toumpas, Kipouros, Papadopoulos, Bekiaridou, Barmpagiannos, Vasilopoulou, Barmpagiannos, Samaras, Giannakoulas, and Tsoumakas}]{BioASQ2025ElCardioCC}
D.~Dimitriadis, V.~Patsiou, E.~Stoikopoulou, A.~Toumpas, A.~Kipouros, D.~Papadopoulos, A.~Bekiaridou, K.~Barmpagiannos, A.~Vasilopoulou, A.~Barmpagiannos, A.~Samaras, G.~Giannakoulas, and G.~Tsoumakas. 2025.
\newblock {Overview of ElCardioCC Task on Clinical Coding in Cardiology at BioASQ 2025}.
\newblock In \emph{CLEF 2025 Working Notes}.

\bibitem[{Garc{\'\i}a-Santa et~al.(2020)Garc{\'\i}a-Santa, Cetina, Cappellato, Eickhoff, Ferro, and Nev{\'e}ol}]{garcia2020fle}
Nuria Garc{\'\i}a-Santa, Kendrick Cetina, L~Cappellato, C~Eickhoff, N~Ferro, and A~Nev{\'e}ol. 2020.
\newblock \href {https://ceur-ws.org/Vol-2696/paper_111.pdf} {Fle at clef ehealth 2020: Text mining and semantic knowledge for automated clinical encoding.}
\newblock In \emph{CLEF (Working Notes)}.

\bibitem[{Henning and C.(2020)}]{Henning2020Multilingual}
Sch{\" a}fer Henning and Friedrich C. 2020.
\newblock \href {https://ceur-ws.org/Vol-2696/paper_212.pdf} {Multilingual {ICD}-10 {Code} {Assignment} with {Transformer} {Architectures} using {MIMIC}-{III} {Discharge} {Summaries}}.
\newblock \emph{Conference and Labs of the Evaluation Forum}.

\bibitem[{Karlin and Amin(2025)}]{karlin2025prompt}
Katarina Karlin and Diana Amin. 2025.
\newblock \href {https://www.diva-portal.org/smash/get/diva2:1959279/FULLTEXT01.pdf} {From prompt to icd-10: Evaluating prompt-based decoder large language models for icd-10 coding using a swedish clinical dataset}.

\bibitem[{Li et~al.(2024)Li, Wang, and Yu}]{li2024exploring}
Rumeng Li, Xun Wang, and Hong Yu. 2024.
\newblock Exploring llm multi-agents for icd coding.
\newblock \emph{arXiv preprint arXiv:2406.15363}.

\bibitem[{L{\'o}pez-Garc{\'\i}a et~al.(2020)L{\'o}pez-Garc{\'\i}a, Jerez, and Veredas}]{lopez2020icb}
Guillermo L{\'o}pez-Garc{\'\i}a, Jos{\'e}~M Jerez, and Francisco~J Veredas. 2020.
\newblock \href {https://ceur-ws.org/Vol-2696/paper_101.pdf} {Icb-uma at clef e-health 2020 task 1: Automatic icd-10 coding in spanish with bert}.
\newblock In \emph{CLEF (Working Notes)}.

\bibitem[{Maatouk(2025)}]{maatouk2025leveraging}
Ola Maatouk. 2025.
\newblock \href {https://www.diva-portal.org/smash/get/diva2:1933618/FULLTEXT02.pdf} {Leveraging llms for icd coding and uncertainty estimation: Can the model's awareness of the hierarchical structureof icd-10 codes impact its prediction performance?}

\bibitem[{Miftahutdinov and Tutubalina(2018)}]{miftahutdinov2018deep}
Zulfat Miftahutdinov and Elena Tutubalina. 2018.
\newblock Deep learning for icd coding: Looking for medical concepts in clinical documents in english and in french.
\newblock In \emph{Experimental IR Meets Multilinguality, Multimodality, and Interaction: 9th International Conference of the CLEF Association, CLEF 2018, Avignon, France, September 10-14, 2018, Proceedings 9}, pages 203--215. Springer.

\bibitem[{Miranda-Escalada et~al.(2020)Miranda-Escalada, Gonzalez-Agirre, Armengol-Estapé, and Krallinger}]{CLEFeHealth2020Task1Overview}
Antonio Miranda-Escalada, Aitor Gonzalez-Agirre, Jordi Armengol-Estapé, and Martin Krallinger. 2020.
\newblock Overview of automatic clinical coding: annotations, guidelines, and solutions for non-english clinical cases at codiesp track of {CLEF eHealth} 2020.
\newblock In \emph{{Working Notes of Conference and Labs of the Evaluation (CLEF) Forum}}, {CEUR} Workshop Proceedings.

\bibitem[{Mustafa et~al.(2025)Mustafa, Naseem, and Azghadi}]{mustafa2025large}
Akram Mustafa, Usman Naseem, and Mostafa~Rahimi Azghadi. 2025.
\newblock Large language models vs human for classifying clinical documents.
\newblock \emph{International Journal of Medical Informatics}, page 105800.

\bibitem[{Pathak et~al.(2024)Pathak, Vald, Sermet, and Demir}]{pathak2024utilizing}
Rudransh Pathak, Gabriel Vald, Yusuf Sermet, and Ibrahim Demir. 2024.
\newblock Utilizing large language models to predict icd-10 diagnosis codes from patient medical records.
\newblock In \emph{2024 IEEE MIT Undergraduate Research Technology Conference (URTC)}, pages 1--5. IEEE.

\bibitem[{Puts et~al.(2025)Puts, Zegers, Dekker, and Bermejo}]{puts2025developing}
Sander Puts, Catharina~ML Zegers, Andre Dekker, and I{\~n}igo Bermejo. 2025.
\newblock Developing an icd-10 coding assistant: Pilot study using roberta and gpt-4 for term extraction and description-based code selection.
\newblock \emph{JMIR Formative Research}, 9:e60095.

\bibitem[{Seva et~al.(2017)Seva, Kittner, Roller, and Leser}]{seva2017multi}
Jurica Seva, Madeleine Kittner, Roland Roller, and Ulf Leser. 2017.
\newblock \href {https://ceur-ws.org/Vol-1866/paper_70.pdf} {Multi-lingual icd-10 coding using a hybrid rule-based and supervised classification approach at clef ehealth 2017.}
\newblock In \emph{CLEF (Working Notes)}.

\bibitem[{Sheng-Wei et~al.(2014)Sheng-Wei, Po-Ting, Yi-Lin, Jay, S., and Richard}]{Sheng2014NCU}
Chen Sheng-Wei, Lai Po-Ting, Tsai Yi-Lin, Kuan-Chieh~Chung Jay, S.~Hsiao S., and Tzong-Han~Tsai Richard. 2014.
\newblock \href {https://citeseerx.ist.psu.edu/document?repid=rep1&type=pdf&doi=0b5662e8ed5af98c2cad9217ee6eb4a14fd63659} {Ncu {IISR} {System} for {NTCIR}-11 {MedNLP}-2 {Task}}.
\newblock \emph{NTCIR Conference on Evaluation of Information Access Technologies}.

\bibitem[{Soroush et~al.(2023)Soroush, Glicksberg, Zimlichman, Barash, Freeman, Charney, Nadkarni, and Klang}]{soroush2023assessing}
Ali Soroush, Benjamin~S Glicksberg, Eyal Zimlichman, Yiftach Barash, Robert Freeman, Alexander~W Charney, Girish~N Nadkarni, and Eyal Klang. 2023.
\newblock Assessing gpt-3.5 and gpt-4 in generating international classification of diseases billing codes.
\newblock \emph{medRxiv}, pages 2023--07.

\bibitem[{Zweigenbaum and Lavergne(2016)}]{Zweigenbaum2016Hybrid}
Pierre Zweigenbaum and Thomas Lavergne. 2016.
\newblock \href {https://doi.org/10.18653/v1/w16-6113} {Hybrid methods for {ICD}-10 coding of death certificates}.
\newblock In \emph{Proceedings of the {Seventh} {International} {Workshop} on {Health} {Text} {Mining} and {Information} {Analysis}}, pages 96--105. Association for Computational Linguistics.

\end{thebibliography}

\appendix
\begin{appendices}

\section*{Appendix A} \label{appendix_a}
\begin{figure} [h!]
   \begin{tcolorbox}[skin=widget,
        boxrule=1mm,
        coltitle=black,
        colframe=blue!30!white,
        colback=blue!2!white,
        width=(.999\textwidth),before=\hfill,after=\hfill,
        adjusted title={ICD-10 Code Generation Prompt}]
       \small{     
        \begin{verbatim} 
You are an expert medical coder, fluent in Greek and possessing an encyclopedic 
knowledge of the ICD-10 coding system. You are meticulous, precise, and 
understand the nuances of medical terminology. Your task is to analyze a patient 
discharge summary written in Greek, identify the medical terms marked with 
asterisks (*term*), and assign the most accurate ICD-10 code to each.

Your output should be a JSON array, where each element in the array is a JSON 
object representing a single medical term and its corresponding ICD-10 code. 

The JSON object should have the following structure:
```json
[
  {
    "medical_term_greek": "*Greek medical term 1*",
    "icd10_code": "ICD-10 Code 1",
    "explanation": "Brief explanation of why this code was chosen, including 
    any assumptions made."
  },
  {
    "medical_term_greek": "*Greek medical term 2*",
    "icd10_code": "ICD-10 Code 2",
    "explanation": "Brief explanation of why this code was chosen, including 
    any assumptions made."
  },
]
```
**Important Considerations:**
*   **Specificity:** Choose the most specific ICD-10 code available.
*   **Context:** Consider the context of the discharge summary when assigning 
codes.
*   **Assumptions:** If you need to make any assumptions to assign a code, 
clearly state them in the "explanation" field.
*   **Multiple Codes:** If a single medical term requires multiple ICD-10 
codes for complete accuracy (e.g., for etiology and manifestation), 
include all relevant codes in the "icd10_code" field, separated by commas.  
Explain the use of multiple codes in the "explanation" field.
*   **Uncertainty:** If you are uncertain about the correct code, provide the 
most likely code and explain your reasoning and any alternative codes you 
considered in the "explanation" field.
*   **Exactness** Make sure to provide a code for all medical terms and 
return the medical term exactly as it appears in the text without changing it
**Example Input:**
``` 
{example_input} 
```
**Example Output:**
```json 
{example_output} 
```
Now, analyze the following patient discharge summary (in Greek) and provide the  
JSON output as described above: {clinical_text}
\end{verbatim}
}
   \end{tcolorbox} 
   \caption{Example prompt for ICD-10 coding in Greek.}\label{figPromptExample}
\end{figure}

\clearpage
\section*{Appendix B} \label{appendix_b}
\begin{table}[h!]
\begin{center}
\begin{tabular}{ p{14cm} } 
\toprule
\textbf{Example Discharge Summary Excerpt} \\
\midrule
La serología hidatídica resultó negativa, por lo que se decidió someter a la paciente a una RMN que tampoco aclaró el diagnóstico de forma definitiva: masa en el polo superior del riñón izquierdo con estructura interna compleja, bien delimitada, sin captación de contraste, con calcificaciones, sospechosa de \textit{\textbf{quiste hidatídico renal}} (correct: B67.99, predicted: B67.4) o \textit{\textbf{nefroma}} (correct: C64.9, predicted: C64.9) quístico multilocular.\\
\textit{(Hydatid serology was negative, so it was decided to subject the patient to an MRI, which also did not definitively clarify the diagnosis: a mass in the upper pole of the left kidney with a complex internal structure, well-defined, without contrast uptake, with calcifications, suspicious for a \textit{\textbf{renal hydatid cyst}} or multilocular cystic \textit{\textbf{nephroma}}.)}\\
\bottomrule
\end{tabular}
\end{center}
\caption{An example excerpt from CodiEsp with two entities with correct ICD-10 codes and their predictions from GPT-4.1 in parenthesis after each entity.}
\label{tab:example_linking}
\end{table}

\end{appendices}

\end{document}